\documentclass{article}
\PassOptionsToPackage{sort&compress}{natbib}

\usepackage[preprint]{neurips_2019}
\usepackage{times}
\usepackage{todonotes}

\usepackage{amsmath,amsfonts,bm}

\newcommand{\figleft}{{\em (Left)}}
\newcommand{\figcenter}{{\em (Center)}}
\newcommand{\figright}{{\em (Right)}}

\def\eqref#1{equation~\ref{#1}}

\def\1{\bm{1}}

\DeclareMathAlphabet{\mathsfit}{\encodingdefault}{\sfdefault}{m}{sl}
\SetMathAlphabet{\mathsfit}{bold}{\encodingdefault}{\sfdefault}{bx}{n}

\newcommand{\E}{\mathbb{E}}

\usepackage{xcolor}
\usepackage{hyperref}
\usepackage{url}
\usepackage{microtype}
\usepackage{graphicx}
\usepackage{soul}
\usepackage{booktabs}
\usepackage{amsmath}
\usepackage{amssymb}
\usepackage{bbm}
\usepackage{wrapfig}
\usepackage{subcaption}
\usepackage{algorithm}
\usepackage{caption}
\usepackage[noend]{algpseudocode}
\usepackage{pifont}
\newcommand{\cmark}{\ding{51}}
\newcommand{\xmark}{\ding{55}}
\newcommand{\rulesep}{\unskip\ \vrule\ }
\usepackage{caption} 
\newcommand{\T}{\mathcal{T}}

\algrenewcommand\algorithmicthen{}

\title{Search on the Replay Buffer:\\
{\Large Bridging Planning and Reinforcement Learning}}

\author{
Benjamin Eysenbach$^{\theta\phi}$, Ruslan Salakhutdinov$^{\theta}$, Sergey Levine$^{\phi\psi}$ \\
$^\theta$CMU, $^\phi$Google Brain, $^\psi$UC Berkeley\\
\texttt{beysenba@cs.cmu.edu} \\
}

\begin{document}

\maketitle

\begin{abstract}
The history of learning for control has been an exciting back and forth between two broad classes of algorithms: planning and reinforcement learning. Planning algorithms effectively reason over long horizons, but assume access to a local policy and distance metric over collision-free paths. Reinforcement learning excels at learning policies and the relative values of states, but fails to plan over long horizons.
Despite the successes of each method in various domains, tasks that require reasoning over long horizons with limited feedback and high-dimensional observations remain exceedingly challenging for both planning and reinforcement learning algorithms.
Frustratingly, these sorts of tasks are potentially the most useful, as they are simple to design (a human only need to provide an example goal state) and avoid reward shaping, which can bias the agent towards finding a sub-optimal solution.
We introduce a general-purpose control algorithm that combines the strengths of planning and reinforcement learning to effectively solve these tasks.
Our aim is to decompose the task of reaching a distant goal state into a sequence of easier tasks, each of which corresponds to reaching a particular subgoal. Planning algorithms can automatically find these waypoints, but only if provided with suitable abstractions of the environment -- namely, a graph consisting of nodes and edges. Our main insight is that this graph can be constructed via reinforcement learning, where a goal-conditioned value function provides edge weights, and nodes are taken to be previously seen observations in a replay buffer.
Using graph search over our replay buffer, we can automatically generate this sequence of subgoals, even in image-based environments. Our algorithm, search on the replay buffer (SoRB), enables agents to solve sparse reward tasks over one hundred steps, and generalizes substantially better than standard RL algorithms.\footnote{Run our algorithm in your browser: \url{http://bit.ly/rl_search}}

\end{abstract}

\section{Introduction}

\label{sec:introduction}

How can agents learn to solve complex, temporally extended tasks?
Classically, planning algorithms give us one tool for learning such tasks. While planning algorithms work well for tasks where it is easy to determine distances between states and easy to design a local policy to reach nearby states, both of these requirements become roadblocks when applying planning to high-dimensional (e.g., image-based) tasks.
Learning algorithms excel at handling high-dimensional observations, but reinforcement learning (RL) -- learning for control -- fails to reason over long horizons to solve temporally extended tasks. 
In this paper, we propose a method that combines the strengths of planning and RL, resulting in an algorithm that can plan over long horizons in tasks with high-dimensional observations.

Recent work has introduced goal-conditioned RL algorithms~\citep{schaul2015universal,pong2018temporal} that acquire a single policy for reaching many goals.
In practice, goal-conditioned RL succeeds at reaching nearby goals but fails to reach distant goals; performance degrades quickly as the number of steps to the goal increases~\citep{nachum2018data,levy2018hierarchical}.
Moreover, goal-conditioned RL often requires large amounts of reward shaping~\citep{autorl} or human demonstrations~\citep{nair2018overcoming,lynch2019learning}, both of which can limit the asymptotic performance of the policy by discouraging the policy from seeking novel solutions.

\begin{figure}[t]
    \centering
    \includegraphics[width=\linewidth]{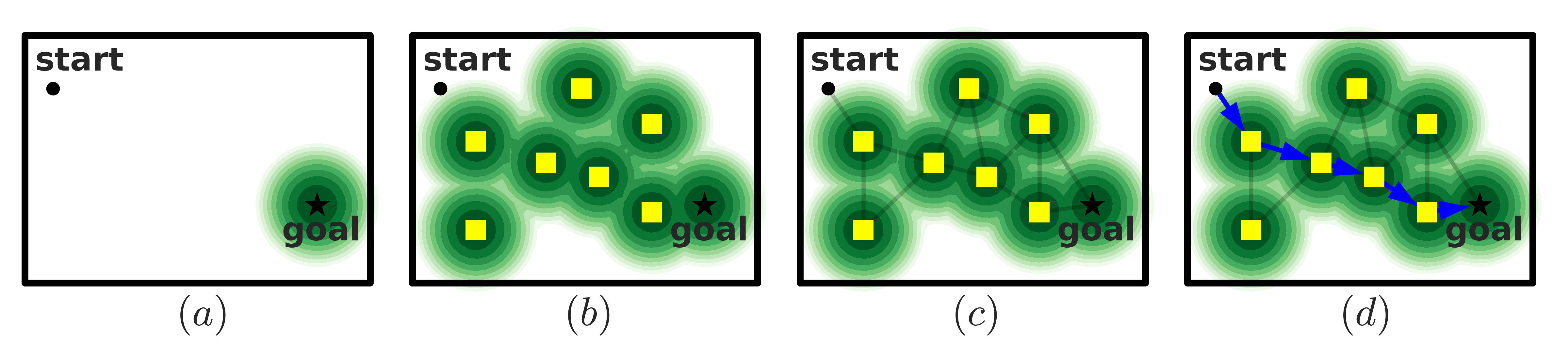}
    \vspace{-1em}
    \caption{\textbf{Search on the Replay Buffer:}
    (a) Goal-conditioned RL often fails to reach distant goals, but can successfully reach the goal if starting nearby (inside the green region). (b) Our goal is to use observations in our replay buffer (yellow squares) as waypoints leading to the goal. (c) We automatically find these waypoints by using the agent's value function to predict when two states are nearby, and building the corresponding graph. (d) We run graph search to find the sequence of waypoints (blue arrows), and then use our goal-conditioned policy to reach each waypoint.}
    \label{fig:teaser}
    \vspace{-1em}
\end{figure}
We propose to solve long-horizon, sparse reward tasks by decomposing the task into a series of easier goal-reaching tasks.
We learn a goal-conditioned policy for solving each of the goal-reaching tasks. Our main idea is to reduce the problem of finding these subgoals to solving a shortest path problem over states that we have previous visited, using a distance metric extracted from our goal-conditioned policy. We call this algorithm Search on Replay Buffer (SoRB), and provide a simple illustration of the algorithm in Figure~\ref{fig:teaser}.

Our primary contribution is an algorithm that bridges planning and deep RL for solving long-horizon, sparse reward tasks. We develop a practical instantiation of this algorithm using ensembles of distributional value functions, which allows us to \emph{robustly} learn distances and use them for \emph{risk-aware} planning.
Empirically, we find that our method generates effective plans to solve long horizon navigation tasks, even in image-based domains, without a map and without odometry.
Comparisons with state-of-the-art RL methods show that SoRB is substantially more successful in reaching distant goals. We also observe that the learned policy generalizes well to navigate in unseen environments.
In summary, graph search over previously visited states is a simple tool for boosting the performance of a goal-conditioned RL algorithm.

\section{Bridging Planning and Reinforcement Learning}
\label{sec:method}

Planning algorithms must be able to (1) sample valid states, (2) estimate the distance between reachable pairs of states, and (3) use a local policy to navigate between nearby states. These requirements are difficult to satisfy in complex tasks with high dimensional observations, such as images. For example, consider a robot arm stacking blocks using image observations. Sampling states requires generating photo-realistic images, and estimating distances and choosing actions requires reasoning about dozens of interactions between blocks. Our method will obtain distance estimates and a local policy using a RL algorithm. To sample states, we will simply use a replay buffer of previously visited states as a non-parametric generative model.

\subsection{Building Block: Goal-Conditioned RL}
\label{sec:prelim}
A key building block of our method is a goal-conditioned policy and its associated value function. We consider a goal-reaching agent interacting with an environment. The agent observes its current state $s \in \mathcal{S}$ and a goal state $s_g \in \mathcal{S}$. The initial state for each episode is sampled $s_1 \sim \rho(s)$, and dynamics are governed by the distribution $p(s_{t+1} \mid s_t, a_t)$. At every step, the agent samples an action $a \sim \pi(a \mid s, s_g)$ and receives a corresponding reward $r(s, a, s_g)$ that indicates whether the agent has reached the goal. The episode terminates as soon as the agent reaches the goal, or after $T$ steps, whichever occurs first. The agent's task is to maximize its cumulative, \emph{undiscounted}, reward. We use an off-policy algorithm to learn such a policy, as well as its associated goal-conditioned Q-function and value function:
\begin{equation*}
    Q(s, a, s_g) = \E_{\substack{s_1 \sim \rho(s), a_t \sim \pi(a_t \mid s_t, s_g)\\s_{t+1} \sim p(s_{t+1} \mid s_t, a_t)}}\left[\sum_{t=1}^T r(s_t, s_g, a_t)\right], \qquad V(s, s_g) = \max_a Q(s, a, s_g)
\end{equation*}
We obtain a policy by acting greedily w.r.t. the Q-function: \mbox{$\pi(a \mid s, s_g) = \arg\max_a Q(s, a, s_g)$}.
We choose an off-policy RL algorithm with goal relabelling~\citep{kaelbling1993learning,andrychowicz2017hindsight} and distributional RL~\citep{bellemare2017distributional}) not only for improved data efficiency, but also to obtain good distance estimates (See Section~\ref{sec:learning-distances}). We will use DQN~\citep{mnih2013playing} for discrete action environments and DDPG~\citep{lillicrap2015continuous} for continuous action environments. Both algorithms operate by minimizing the Bellman error over transitions sampled from a replay buffer $\mathcal{B}$.

\subsection{Distances from Goal-Conditioned Reinforcement Learning}
\label{sec:learning-distances}
To ultimately perform planning, we need to compute the \emph{shortest path distance} between pairs of states. Following~\citet{kaelbling1993learning}, we define a reward function that returns -1 at every step: \mbox{$r(s, a, s_g) \triangleq -1$}.
The episode ends when the agent is sufficiently close to the goal, as determined by a state-identity oracle.
Using this reward function and termination condition, there is a close connection between the Q values and shortest paths.
We define $d_{\text{sp}}(s, s_g)$ to be the shortest path distance from state $s$ to state $s_g$. That is, $d_{\text{sp}}(s, g)$ is the expected number of steps to reach $s_g$ from $s$ under the optimal policy.
The value of state $s$ with respect to goal $s_g$ is simply the negative shortest path distance: $V(s, s_g) = -d_{\text{sp}}(s, s_g)$. We likewise define $d_{\text{sp}}(s, a, s_g)$ as the shortest path distance, conditioned on initially taking action $a$. Then Q values also equal a negative shortest path distance: $Q(s, a, s_g) = -d_{\text{sp}}(s, a, s_g)$.
Thus, goal-conditioned RL on a suitable reward function yields a Q-function that allows us to estimate shortest-path distances.

\subsection{The Replay Buffer as a Graph}
We build a weighted, \emph{directed} graph directly on top of states in our replay buffer, so each node corresponds to an observation (e.g., an image). We add edges between nodes with weight (i.e., length) equal to their predicted distance, but ignore edges that are longer than \textsc{MaxDist}, a hyperparameter:
\begin{align*}
    \mathcal{G} \triangleq (\mathcal{V}, \mathcal{E}, \mathcal{W}) \qquad \text{where} 
    \quad & \mathcal{V} = \mathcal{B}, \quad \mathcal{E} = \mathcal{B} \times \mathcal{B} = \{e_{s_1 \rightarrow s_2} \mid s_1, s_2 \in \mathcal{B} \} \\
          &\mathcal{W}(e_{s_1 \rightarrow s_2}) = \begin{cases} d_\pi(s_1, d_2) & \text{if } d_\pi(s_1, s_2) < \textsc{MaxDist} \\
          \infty & \text{otherwise} \end{cases}
\end{align*}
Given a start and goal state, we temporarily add each to the graph. We add directed edges from the start state to every other state, and from every other state to the goal state, using the same criteria as above.
We use Dijkstra's Algorithm to find the shortest path. See Appendix~\ref{appendix:graph-search} for details.

\subsection{Algorithm Summary}

\begin{wrapfigure}[16]{R}{0.5\textwidth}
\vspace{-2em}
\begin{minipage}[t]{0.5\textwidth}
  \begin{algorithm}[H]
    \caption{Inputs are the current state $s$, the goal state $s_g$, a buffer of observations $\mathcal{B}$, the learned policy $\pi$ and its value function $V$. Returns an action $a$.
    \label{alg:search-policy} 
   }
    \begin{algorithmic}
    \Function{SearchPolicy}{$s, s_g, \mathcal{B}, V, \pi$}
   \State $s_{w_1}, \cdots \gets \Call{ShortestPath}{s, s_g, \mathcal{B}, V}$
   \State $d_{s \rightarrow w_1} \gets -V(s, s_{w_1})$
   \State $d_{s \rightarrow g} \gets -V(s, s_g)$
   \If{$d_{s \rightarrow w_1} < d_{s \rightarrow g}$ or $d_{s \rightarrow g} > $ \textsc{MaxDist}}
      \State $a \gets \pi(a, \mid s, {\color{blue}s_{w_1}})$
   \Else{}
      \State $a \gets \pi(a, \mid s, {\color{blue}s_g})$
   \EndIf
   \State \Return $a$
   \EndFunction
    \end{algorithmic}
  \end{algorithm}
\end{minipage}
\end{wrapfigure}
After learning a goal-conditioned Q-function, we perform graph search to find a set of waypoints and use the goal-conditioned policy to reach each. We view the combination of graph search and the underlying goal-conditioned policy as a new \textsc{SearchPolicy}, shown in Algorithm~\ref{alg:search-policy}. The algorithm starts by using graph search to obtain the shortest path $s_{w_1}, s_{w_2}, \cdots$ from the current state $s$ to the goal state $s_g$, planning over the states in our replay buffer $\mathcal{B}$.
We then estimate the distance from the current state to the first waypoint, as well as the distance from the current state to the goal.
In most cases, we then condition the policy on the first waypoint, $s_{w_1}$. However, if the goal state is closer than the next waypoint and the goal state is not too far away, then we directly condition the policy on the final goal.
If the replay buffer is empty or there is not a path in $\mathcal{G}$ to the goal, then Algorithm~\ref{alg:search-policy} resorts to standard goal-conditioned RL.

\section{Better Distance Estimates}
The success of our \textsc{SearchPolicy} depends heavily on the accuracy of our distance estimates. This section proposes two techniques to learn better distances with RL.
\vspace{-0.2em}

\subsection{Better Distances via Distributional Reinforcement Learning}

\begin{wrapfigure}[18]{R}{0.5\textwidth}
    \vspace{-1.5em}
    \centering
    \includegraphics[width=\linewidth]{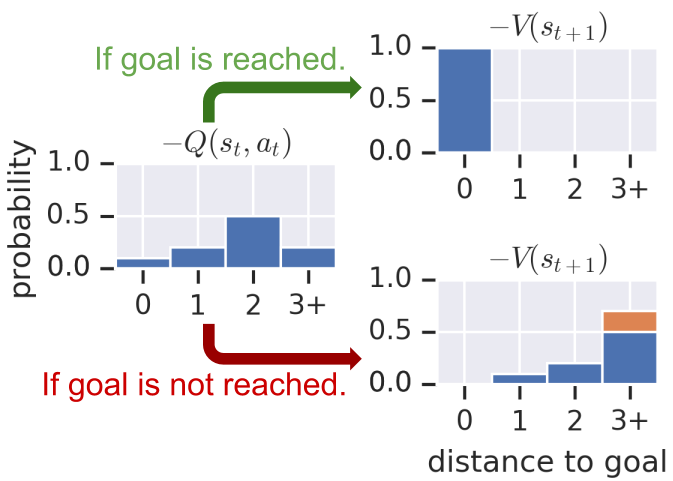}
    \vspace{-1em}
    \caption{The Bellman update for distributional RL is simple when learning distances, simply corresponding to a left-shift of the Q-values at every step until the agent reaches the goal.}
    \label{fig:q-vals}
\end{wrapfigure}
Off-the-shelf Q-learning algorithms such as DQN~\citep{mnih2013playing} or DDPG~\citep{lillicrap2015continuous} will fail to learn accurate distance estimates using the $-1$ reward function. The true value for a state and goal that are unreachable is $-\infty$, which cannot be represented by a standard, feed-forward Q-network. Simply clipping the Q-value estimates to be within some range avoids the problem of ill-defined Q-values, but empirically we found it challenging to train clipped Q-networks.
We adopt distributional Q-learning~\citep{bellemare2017distributional}, noting that is has a convenient form when used with the $-1$ reward function.
Distributional RL discretizes the possible value estimates into a set of bins \mbox{$B = (B_1, B_2, \cdots, B_N)$}. For learning distances, bins correspond to distances, so $B_i$ indicates the event that the current state and goal are $i$ steps away from one another. Our Q-function predicts a distribution $Q(s_t, s_g, a_t) \in \mathcal{P}^N$ over these bins, where $Q(s_t, s_g, a_t)_i$ is the predicted probability that states $s_t$ and $s_g$ are $i$ steps away from one another. To avoid ill-defined Q-values, the final bin, $B_N$ is a catch-all for predicted distances of at least $N$. Importantly, this gives us a well-defined method to represent large and infinite distances. Under this formulation, the targets $Q^* \in \mathcal{P}^N$ for our Q-values have a simple form:
\begin{equation*}
    Q^* = \begin{cases}
    (1, 0, \cdots, 0) & \text{if } s_t = g \\
    (0, Q_1, \cdots, Q_{N-2}, Q_{N-1} + Q_{N}) & \text{if } s_t \neq g 
    \end{cases}
\end{equation*}
As illustrated in Figure~\ref{fig:q-vals}, if the state and goal are equivalent, then the target places all probability mass in bin 0. Otherwise, the targets are a right-shift of the current predictions. To ensure the target values sum to one, the mass in bin $N$ of the targets is the sum of bins $N-1$ and $N$ from the predicted values. Following~\citet{bellemare2017distributional}, we update our Q function by minimizing the KL divergence between our predictions $Q^\theta$ and the target $Q^*$:
\begin{equation}
    \min_\theta D_{\text{KL}}(Q^* \; \| \; Q^\theta) \label{eq:kl}
\end{equation}

\subsection{Robust Distances via Ensembles of Value Functions}
\label{sec:ensembles}
Since we ultimately want to use estimated distances to perform search, it is crucial that we have accurate distances estimates.
It is challenging to robustly estimate the distance between all $|\mathcal{B}|^2$ pairs of states in our buffer $\mathcal{B}$, some of which may not have occurred during training. If we fail and spuriously predict that a pair of distant states are nearby, graph search will exploit this ``wormhole'' and yield a path which assumes that the agent can ``teleport'' from one distant state to another.
We seek to use a bootstrap~\citep{bickel1981some} as a principled way to estimate uncertainty for our Q-values. Following prior work~\citep{osband2016deep,lakshminarayanan2017simple}, we implement an approximation to the bootstrap. We train an ensemble of Q-networks, each with independent weights, but trained on the same data using the same loss (Eq.~\ref{eq:kl}).
When performing graph search, we aggregate predictions from each Q-network in our ensemble. Empirically, we found that ensembles were crucial for getting graph search to work on image-based tasks, but we observed little difference in whether we took the maximum predicted distance or the average predicted distance.

\vspace{-0.5em}
\section{Related Work}
\label{sec:related-work}
\vspace{-0.5em}

\emph{Planning Algorithms}:
Planning algorithms~\citep{lavalle2006planning, choset2005principles} efficiently solve long-horizon tasks, including those that stymie RL algorithms (see, e.g.,~\citet{levine2011space,kavraki1996probabilistic, lau2005behavior}).
However, these techniques assume that we can (1) efficiently sample valid states, (2) estimate the distance between two states, and (3) acquire a local policy for reaching nearby states, all of which make it challenging to apply these techniques to high-dimensional tasks (e.g., with image observations).
Our method removes these  assumptions by (1) sampling states from the replay buffer and (2,3) learning the distance metric and policy with RL.
Some prior works have also combined planning algorithms with RL~\citep{autorl,prm-rl, savinov2018semi}, finding that the combination yields agents adept at reaching distant goals. Perhaps the most similar work is Semi-Parametric Topological Memory~\citep{savinov2018semi}, which also uses graph search to find waypoints for a learned policy. We compare to SPTM in Section~\ref{sec:sptm}.

\emph{Goal-Conditioned RL}:
Goal-conditioned policies~\citep{kaelbling1993learning,schaul2015universal,pong2018temporal} take as input the current state and a goal state, and predict a sequence of actions to arrive at the goal.
Our algorithm learns a goal-conditioned policy to reach waypoints along the planned path.
Recent algorithms~\citep{andrychowicz2017hindsight,pong2018temporal} combine off-policy RL algorithms with goal-relabelling to improve the sample complexity and robustness of goal-conditioned policies.
Similar algorithms have been proposed for visual navigation~\citep{anderson2018vision, gupta2017cognitive, zhu2017target, mirowski2016learning}.
A common theme in recent work is learning distance metrics to accelerate RL. While most methods~\citep{florensa2019self,savinov2018episodic,wu2018laplacian} simply perform RL on top of the learned representation, our method explicitly performs search using the learned metric.

\emph{Hierarchical RL}:
Hierarchical RL algorithms automatically learn a set of primitive skills to help an agent learn complex tasks. One class of methods~\citep{kaelbling1993hierarchical, parr1998reinforcement,sutton1999between, precup2000temporal, vezhnevets2017feudal, nachum2018data, frans2017meta, bacon2017option, kulkarni2016hierarchical} jointly learn a low-level policy for performing each of the skills together with a high-level policy for sequencing these skills to complete a desired task. Another class of algorithms~\citep{fox2017multi, csimcsek2005identifying, drummond2002accelerating} focus solely on automatically discovering these skills or subgoals.
SoRB learns primitive skills that correspond to goal-reaching tasks, similar to~\citet{nachum2018data}. While jointly learning high-level and low-level policies can be unstable (see discussion in~\citet{nachum2018data}), we sidestep the problem by using graph search as a fixed, high-level policy.

\begin{wrapfigure}[11]{R}{0.5\textwidth}
    \vspace{-0.5em}
    \begin{tabular}{c|p{1.0cm}|p{1.0cm}|p{1.6cm}}
        model & real states & multi-step & prediction dimension \\
        \hline
        state-space & \cmark & \cmark & 1000s+ \\
        latent-space & \xmark & \cmark & 10s \\
        inverse & \cmark & \xmark & 10s\\
        SoRB & \cmark & \cmark & 1\\
    \end{tabular}
    \caption{Four classes of model-based RL methods. Dimensions in the last column correspond to typical robotics tasks with image/lidar observations.} 
    \label{fig:mbrl-table}
\end{wrapfigure}
\emph{Model Based RL}:
RL methods are typically divided into model-free~\citep{williams1992simple, schulman2015high, schulman2015trust, schulman2017proximal} and model-based~\citep{watkins1992q, lillicrap2015continuous} approaches. Model-based approaches all perform some degree of planning, from predicting the value of some state~\citep{silver2016mastering,mnih2013playing}, obtaining representations by unrolling a learned dynamics model~\citep{racaniere2017imagination}, or learning a policy directly on a learned dynamics model~\citep{sutton1990integrated, chua2018deep, kurutach2018model, finn2017deep, agrawal2016learning, oh2015action, nagabandi2018neural}. One line of work~\citep{amos2018differentiable, srinivas2018universal, tamar2016value, lee2018gated} embeds a differentiable planner inside a policy, with the planner learned end-to-end with the rest of the policy. Other work~\citep{watter2015embed, lenz2015deepmpc} explicitly learns a representation for use inside a standard planning algorithm.
In contrast, SoRB learns to predict the distances between states, which can be viewed as a high-level inverse model. SoRB predicts a scalar (the distance) rather than actions or observations, making the prediction problem substantially easier. By planning over previously visited states, SoRB does not have to cope with infeasible states that can be predicted by forward models in state-space and latent-space.

\vspace{-0.5em}
\section{Experiments}
\label{sec:experiments}
\vspace{-0.5em}

We compare SoRB to prior methods on two tasks: a simple 2D environment, and then a visual navigation task, where our method will plan over images. Ablation experiments will illustrate that accurate distances estimates are crucial to our algorithm's success.

\begin{figure}[t]
    \vspace{-1em}
    \centering
    \begin{subfigure}[b]{0.13\textwidth}
        \centering
      \includegraphics[width=\linewidth]{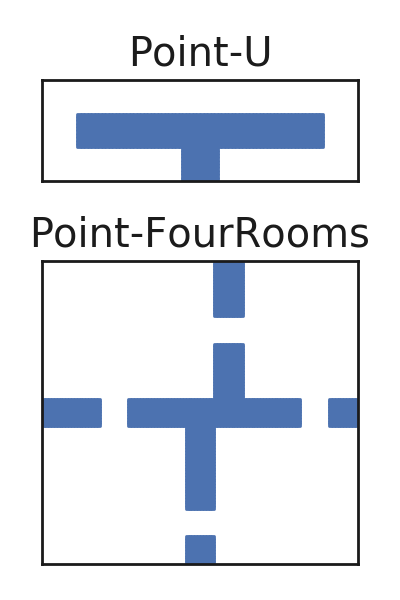}
      \caption{\label{fig:point-env}}
    \end{subfigure}
    \hspace{-1em}
        \begin{subfigure}[b]{0.52\textwidth}
      \centering 
      \includegraphics[width=0.9\linewidth]{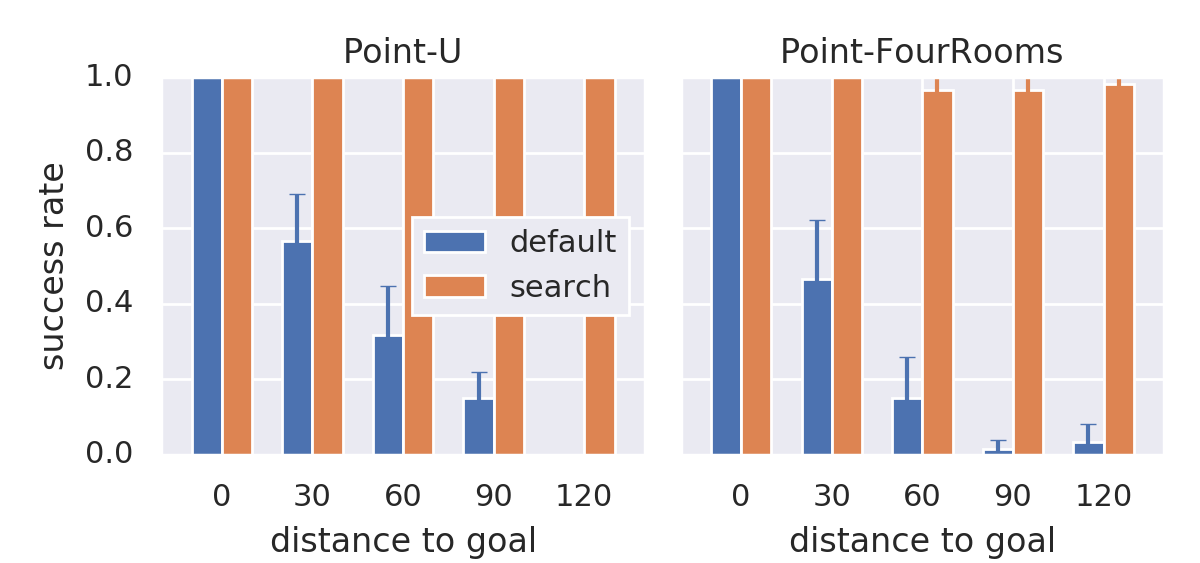}
      \caption{\label{fig:point-results}}
    \end{subfigure}
        \hspace{-1em}
    \begin{subfigure}[b]{0.37\textwidth}
    \includegraphics[width=\linewidth]{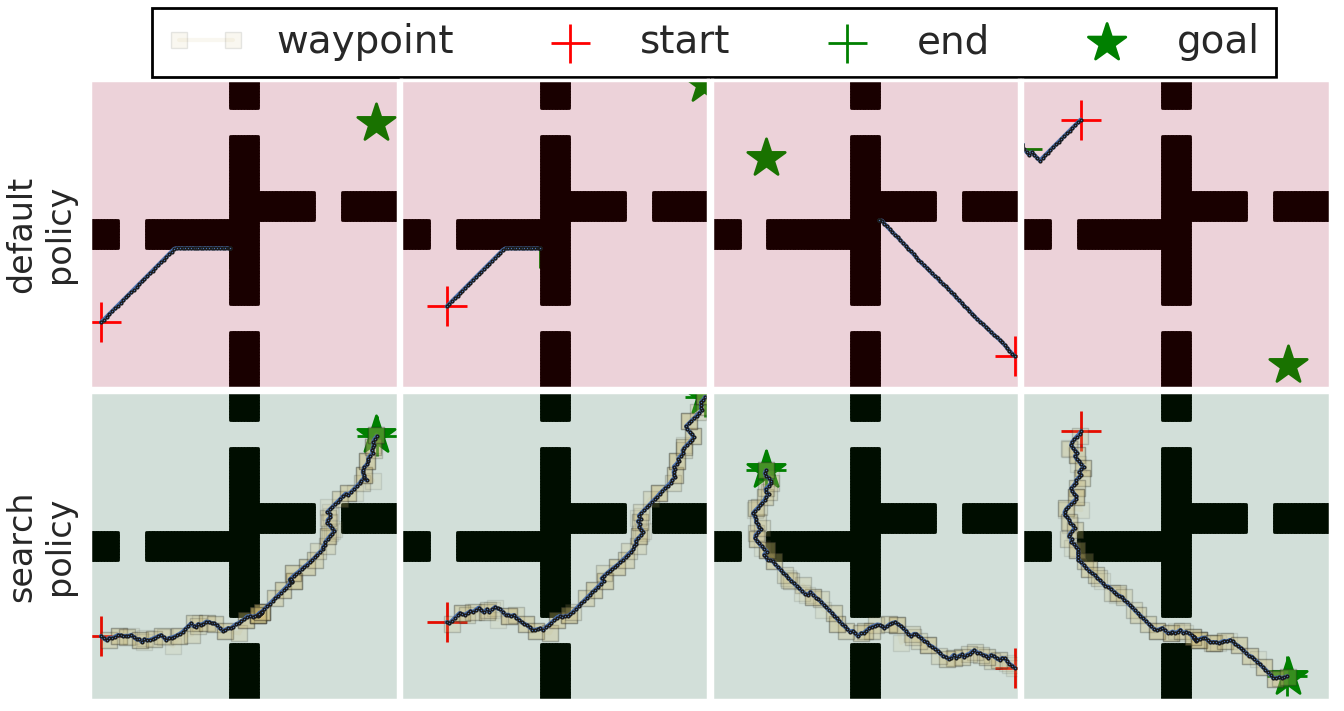}
    \caption{\label{fig:point-rollouts}}
    \end{subfigure}
   \caption{\textbf{Simple 2D Navigation}: \figleft\; Two simple navigation environments. \figcenter\; An agent that combines a goal-conditioned policy with search is substantially more successful at reaching distant goals in these environments than using the goal-conditioned policy alone. \figright\; A standard goal-conditioned policy (top) fails to reach distant goals. Applying graph search on top of that \emph{same policy} (bottom) yields a sequence of intermediate waypoints (yellow squares) that enable the agent to successfully reach distant goals.
   \label{fig:2d}}
   \vspace{-1em}
\end{figure}

\subsection{Didactic Example: 2D Navigation}

We start by building intuition for our method by applying it to two simple 2D navigation tasks, shown in Figure~\ref{fig:point-env}.
The start and goal state are chosen randomly in free space, and reaching the goal often takes over 100 steps, even for the optimal policy.
We used goal-conditioned RL to learn a policy for each environment, and then evaluated this policy on randomly sampled (start, goal) pairs of varying difficulty. To implement SoRB, we used exactly the same policy, both to perform graph search and then to reach each of the planned waypoints.
In Figure~\ref{fig:point-results}, we observe that the goal-conditioned policy can reach nearby goals, but fails to generalize to distant goals. In contrast, SoRB successfully reaches goals over 100 steps away, with little drop in success rate. Figure~\ref{fig:point-rollouts} compares rollouts from the goal-conditioned policy and our policy. Note that our policy takes actions that temporarily lead away from the goal so the agent can maneuver through a hallway to eventually reach the goal.

\subsection{Planning over Images for Visual Navigation}

\begin{wrapfigure}[16]{R}{0.5\textwidth}
    \vspace{-1em}
    \centering
    \includegraphics[width=\linewidth]{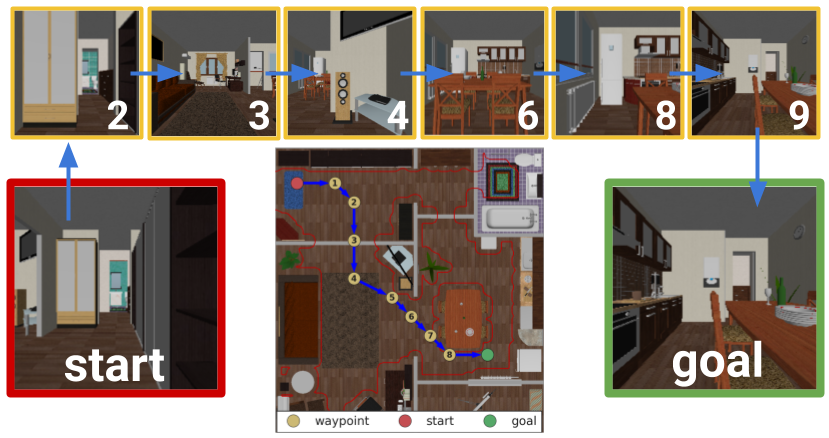}
    \vspace{-.5em}
    \caption{\textbf{Visual Navigation}: Given an initial state and goal state, our method automatically finds a sequence of intermediate waypoints. The agent then follows those waypoints to reach the goal.}
    \vspace{-1em}
    \label{fig:viz-teaser}
\end{wrapfigure}
We now examine how our method scales to high-dimensional observations in a visual navigation task, illustrated in Figure~\ref{fig:viz-teaser}. We use 3D houses from the SUNCG dataset~\citep{song2017semantic}, similar to the task described by~\citet{shah2018follownet}.
The agent receives either RGB or depth images and takes actions to move North/South/East/West. Following~\citet{shah2018follownet}, we stitch four images into a panorama, so the resulting observation has dimension $4 \times 24 \times 32 \times C$, where $C$ is the number of channels (3 for RGB, 1 for Depth).
At the start of each episode, we randomly sample an initial state and goal state. We found that sampling nearby goals (within 4 steps) more often (80\% of the time) improved the performance of goal-conditioned RL. We use the same goal sampling distribution for all methods.
The agent observes both the current image and the goal image, and should take actions that lead to the goal state. The episode terminates once the agent is within 1 meter of the goal. We also terminate if the agent has failed to reach the goal after 20 time steps, but treat the two types of termination differently when computing the TD error (see~\citet{pardo2017time}). 
Note that it is challenging to specify a meaningful distance metric and local policy on pixel inputs, so it is difficult to apply standard planning algorithms to this task.

\begin{figure}[h]
    \centering
    \vspace{-1em}
    \includegraphics[width=\textwidth]{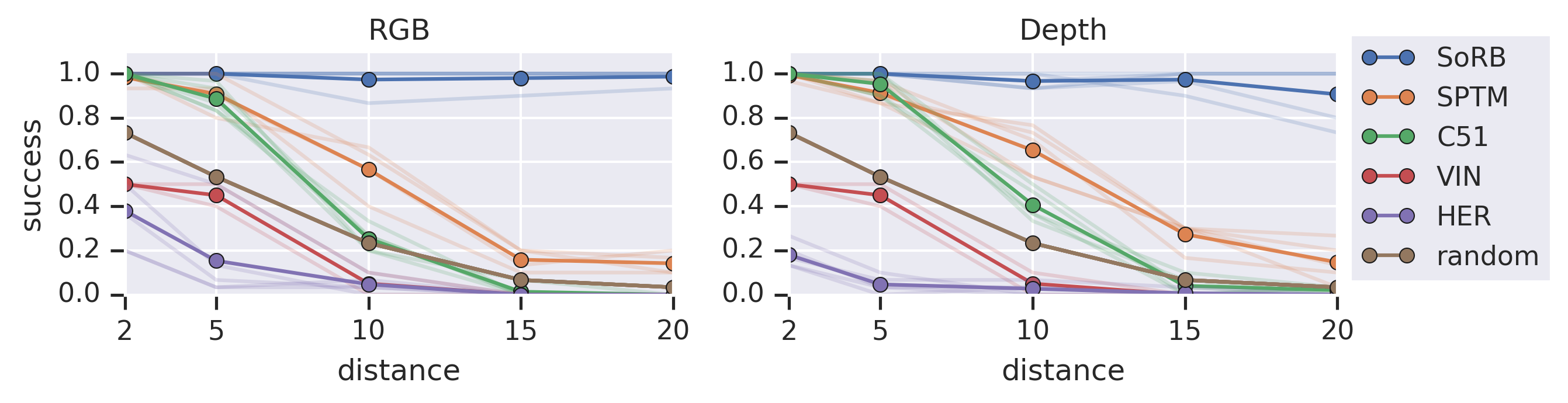}
    \vspace{-1.5em}
    \caption{\textbf{Visual Navigation}: We compare our method (SoRB) to prior work on the visual navigation environment (Fig.~\ref{fig:viz-teaser}), using RGB images \figleft \; and depth images \figright \;. We find that only our method succeeds in reaching distant goals. \emph{Baselines}: SPTM~\citep{savinov2018semi}, C51~\citep{bellemare2017distributional}, VIN~\citep{tamar2016value}, HER~\citep{andrychowicz2017hindsight}.}
    \label{fig:visual-navigation}
\end{figure}

On this task, we evaluate four state-of-the-art prior methods: hindsight experience replay (HER)~\citep{andrychowicz2017hindsight}, distributional RL (C51)~\citep{bellemare2017distributional}, semi-parametric topological memory (SPTM)~\citep{savinov2018semi}, and value iteration networks (VIN)~\citep{tamar2016value}.
SoRB uses C51 as its underlying goal-conditioned policy.
For VIN, we tuned the number of iterations as well as the number of hidden units in the recurrent layer. For SPTM, we performed a grid search over the threshold for adding edges, the threshold for choosing the next waypoint along the shortest path, and the parameters for sampling the training data. In total, we performed over 1000 experiments to tune baselines, more than an order of magnitude more than we used for tuning our own method. See Appendix~\ref{appendix:hparams} for details.

We evaluated each method on goals ranging from 2 to 20 steps from the start. For each distance, we randomly sampled 30 (start, goal) pairs, and recorded the average success rate, defined as reaching within 1 meter of the goal within 100 steps. We then repeated each experiment for 5 random seeds.
In Figure~\ref{fig:visual-navigation}, we plot each random seed as a transparent line; the solid line corresponds to the average across the 5 random seeds. While all prior methods degrade quickly as the distance to the goal increases, our method continues to succeed in reaching goals with probability around 90\%. SPTM, the only prior method that also employs search, performs second best, but substantially worse than our method.

\subsection{Comparison with Semi-Parametric Topological Memory}
\label{sec:sptm}

\begin{figure}[h]
    \centering
    \vspace{-1em}
    \begin{subfigure}[b]{0.59\textwidth}
        \includegraphics[width=\textwidth]{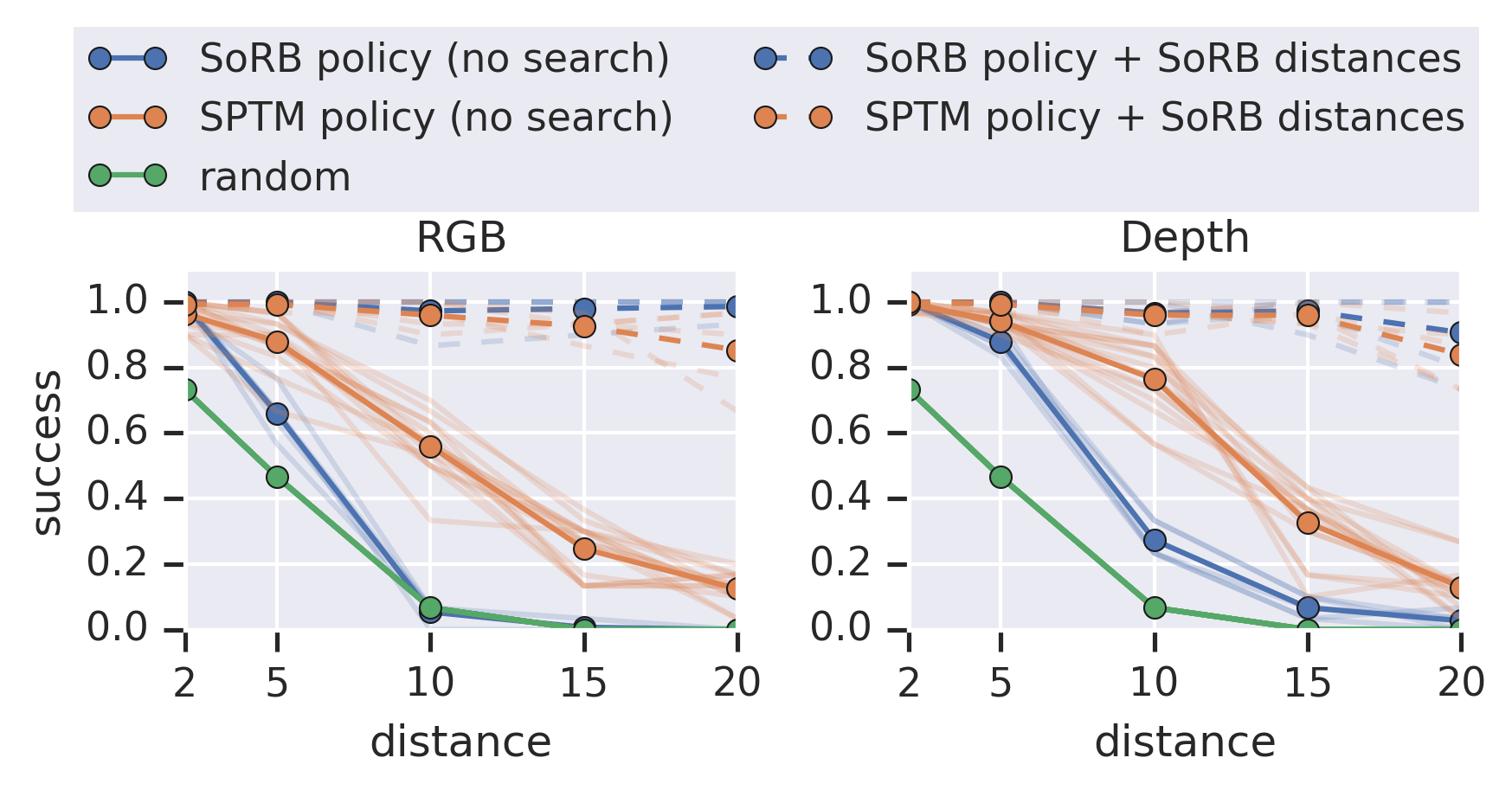}
        \caption{Goal-Conditioned Policy \label{fig:ablation-sptm-policy}}
    \end{subfigure}
    \rulesep
    \begin{subfigure}[b]{0.39\textwidth}
            \includegraphics[width=\textwidth]{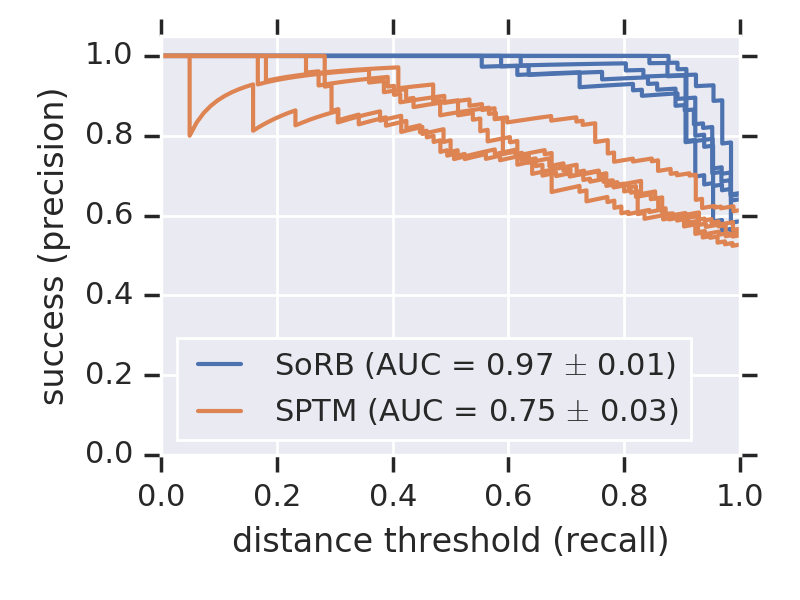}
            \caption{Distance Predictions~\label{fig:dist-pred}}
    \end{subfigure}
\caption{\textbf{SoRB vs SPTM}: Our method and Semi-Parametric Topological Memory~\citep{savinov2018episodic} differ in the policy used and how distances are estimated. We find \figleft \, that both methods learn comparable policies, but \figright\, our method learns more accurate distances. See text for details. \label{fig:ablation-sptm}} \vspace{-0.5em}
\end{figure}

To understand why SoRB succeeds at reaching distant goals more frequently than SPTM, we examine the two key differences between the methods: (1) the \emph{goal-conditioned policy} used to reach nearby goals and (2) the \emph{distance metric} used to construct the graph.
While SoRB acquires a goal-conditioned policy via goal-conditioned RL, SPTM obtains a policy by learning an inverse model with supervised learning. First, we compared the performance of the RL policy (used in SoRB) with the inverse model policy (used in SPTM). 
In Figure~\ref{fig:ablation-sptm-policy}, the solid colored lines show that, \emph{without search}, the policy used by SPTM is more successful than the RL policy, but performance of both policies degrades as the distance to the goal increases. We also evaluate a variant of our method that uses the policy from SPTM to reach each waypoint, and find (dashed-lines) no difference in performance, likely because the policies are equally good at reaching nearby goals (within \textsc{MaxDist} steps). We conclude that the difference in goal-conditioned policies cannot explain the difference in success rate.

The other key difference between SoRB and SPTM is their learned distance metrics.
When using distances for graph search, it is critical for the predicted distance between two states to reflect whether the policy can successfully navigate between those states: the model should be more successful at reaching goals which it predicts are nearby. We can naturally measure this alignment using the area under a precision recall curve. Note that while SoRB predicts distances in the range $[0, T]$, SPTM predicts whether two states are reachable, so its predictions will be in the range $[0, 1]$. Nonetheless, precision-recall curves\footnote{We negate the distance prediction from SoRB before computing the precision recall curve because small distances indicate that the policy should be more successful.} only depend on the ordering of the predictions, not their absolute values. Figure~\ref{fig:dist-pred} shows that the distances predicted by SoRB more accurately reflect whether the policy will reach the goal, as compared with SPTM. The average AUC across five random seeds is 22\% higher for SoRB than SPTM. In retrospect, this finding is not surprising: while SPTM employs a learned, inverse model policy, it learns distances w.r.t. a random policy.

\subsection{Better Distance Estimates}
\begin{figure}[h]
    \vspace{-1em}
    \centering
    \begin{subfigure}[b]{0.5\textwidth}
    \includegraphics[width=\textwidth]{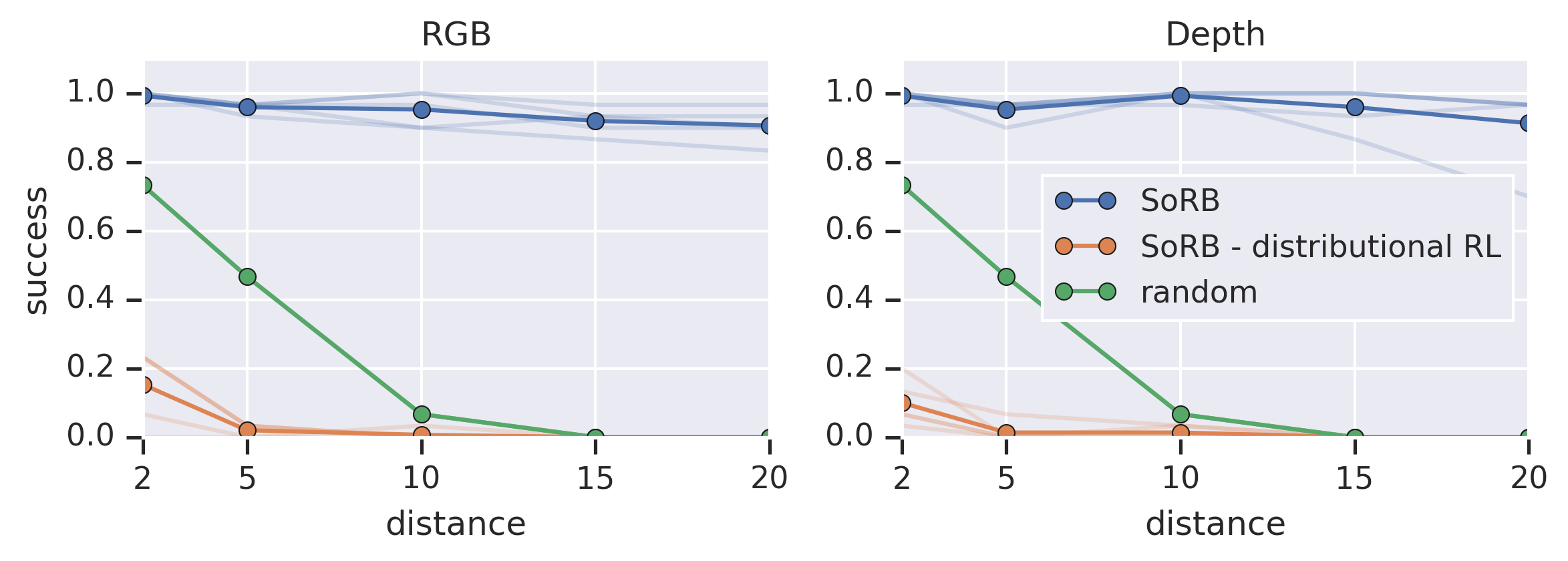}
    \caption{Distributional RL\label{fig:ablation-dist-c51}}
    \end{subfigure}%
    ~
    \begin{subfigure}[b]{0.5\textwidth}
    \includegraphics[width=\textwidth]{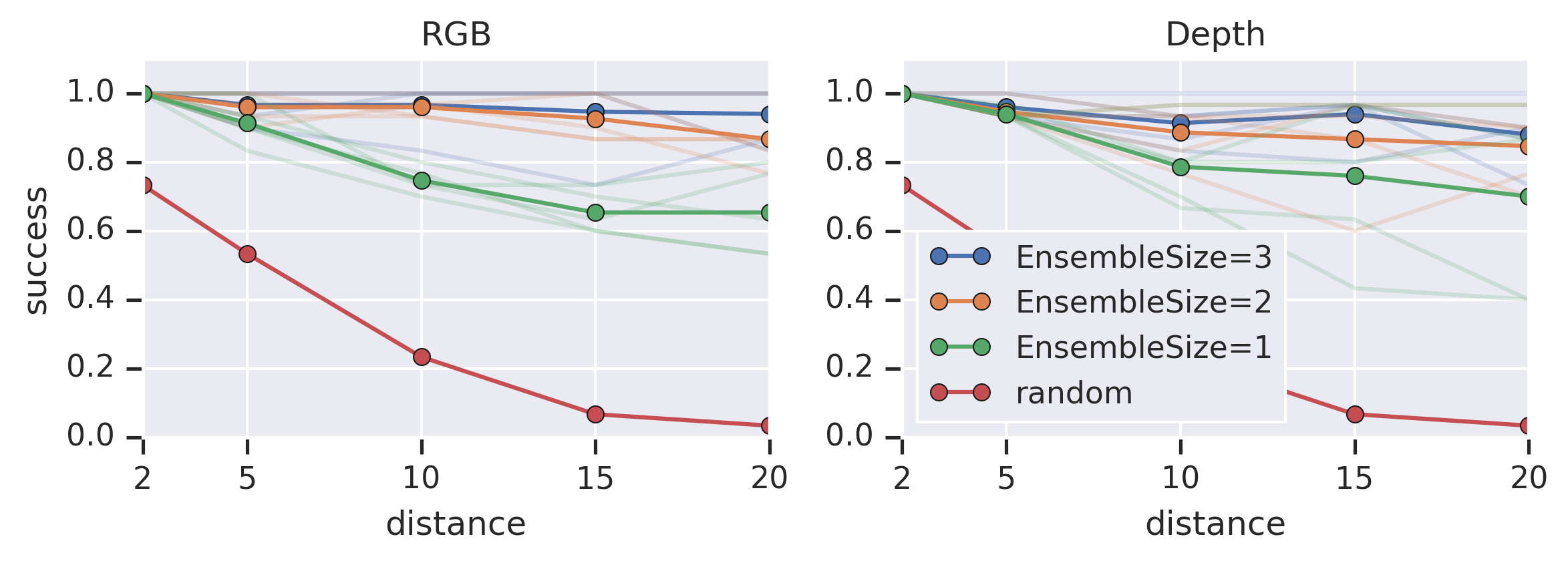}
    \caption{Ensembles\label{fig:ablation-dist-ensembles}}
    \end{subfigure}
    \vspace{-1em}
    \caption{\textbf{Better Distance Estimates}: \figleft \; Without distributional RL, our method performs poorly. \figright \; Ensembles contribute to a moderate increase in success rate, especially for distant goals. \label{fig:ablation-dist}}
    \vspace{-1em}
\end{figure}
We now examine the ingredients in SoRB that contribute to its accurate distance estimates: distributional RL and ensembles of value functions. In a first experiment, evaluated a variant of SoRB trained without distributional RL. As shown in Figure~\ref{fig:ablation-dist-c51}, this variant performed worse than the random policy, clearly illustrating that distributional RL is a key component of SoRB. The second experiment studied the effect of using ensembles of value functions. Recalling that we introduced ensembles to avoid erroneous distance predictions for distant pairs of states, we expect that ensembles will contribute most towards success at reaching distant goals. Figure~\ref{fig:ablation-dist-ensembles} confirms this prediction, illustrating that ensembles provide a 10 - 20\% increase in success at reaching goals that are at least 10 steps away. We run additional ablation analysis in Appendix~\ref{appendix:ablation}.

\subsection{Generalizing to New Houses}
\label{sec:generalization}

\begin{figure}[h]
    \vspace{-1em}
    \centering
    \includegraphics[width=\textwidth]{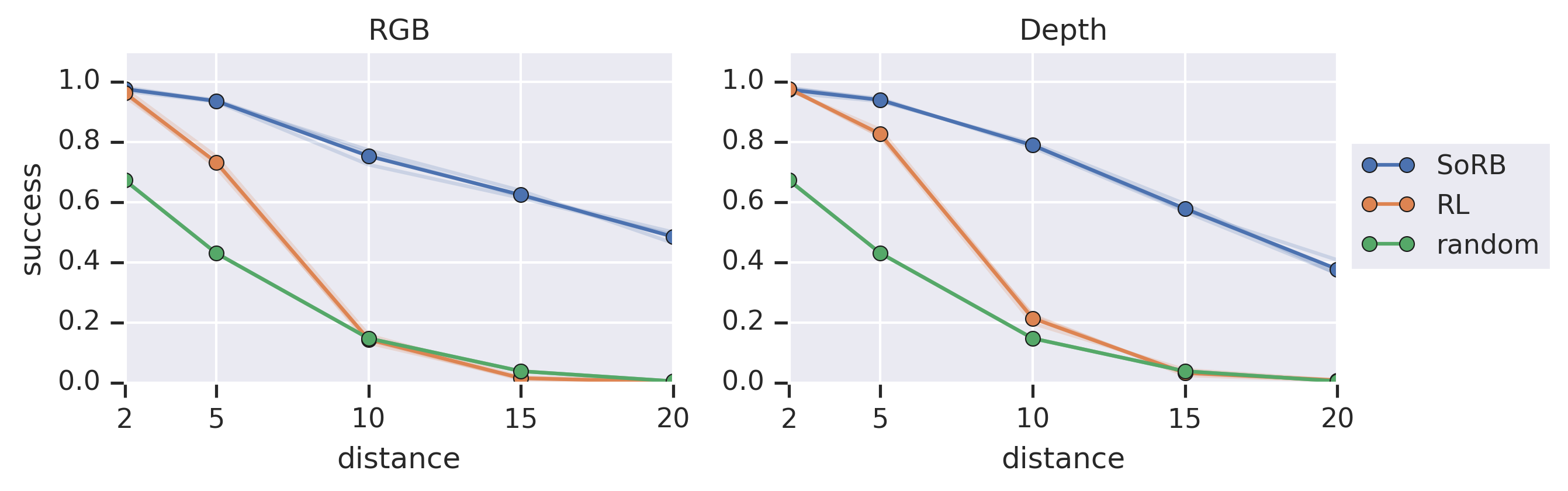}
    \vspace{-1.5em}
    \caption{\textbf{Does SoRB Generalize?} After training on 100 SUNCG houses, we collect random data in held-out houses to use for search in those new environments. Whether using depth images or RGB images, SoRB generalizes well to new houses, reaching almost 80\% of goals 10 steps away, while goal-conditioned RL reaches less than 20\% of these goals. Transparent lines correspond to average success rate across 22 held-out houses for each of three random seeds.}
    \label{fig:generalization}
    \vspace{-0.5em}
\end{figure}
We now study whether our method generalizes to new visual navigation environments. We train on 100 SUNCG houses, randomly sampling one per episode. We evaluated on a held-out test set of 22 SUNCG houses. In each house, we collect 1000 random observations and use those observations to perform search. We use the same goal-conditioned policy and associated distance function that we learned during training. As before, we measure the fraction of goals reached as we increase the distance to the goal.
In Figure~\ref{fig:generalization}, we observe that SoRB reaches almost 80\% of goals that are 10 steps away, about four times more than reached by the goal-conditioned RL agent. Our method succeeds in reaching 40\% of goals 20 steps away, while goal-conditioned RL has a success rate near 0\%. We repeated the experiment for three random seeds, retraining the policy from scratch each time. Note that there is no discernible difference between the three random seeds, plotted as transparent lines, indicating the robustness of our method to random initialization.

\section{Discussion and Future Work}

We presented SoRB, a method that combines planning via graph search and goal-conditioned RL. By exploiting the structure of goal-reaching tasks, we can obtain policies that generalize substantially better than those learned directly from RL. In our experiments, we show that SoRB can solve temporally extended navigation problems, traverse environments with image observations, and generalize to new houses in the SUNCG dataset.
Our method relies heavily on goal-conditioned RL, and we expect advances in this area to make our method applicable to even more difficult tasks.
While we used a stage-wise procedure, first learning the goal-conditioned policy and then applying graph search, in future work we aim to explore how graph search can improve the goal-conditioned policy itself, perhaps via policy distillation or obtaining better Q-value estimates.
In addition, while the planning algorithm we use is simple (namely, Dijkstra), we believe that the key idea of using distance estimates obtained from RL algorithms for planning will open doors to incorporating more sophisticated planning techniques into RL.

\vspace{2em}
{\footnotesize
    \textbf{Acknowledgements}: We thank Vitchyr Pong, Xingyu Lin, and Shane Gu for helpful discussions on learning goal-conditioned value functions, Aleksandra Faust and Brian Okorn for feedback on connections to planning, and Nikolay Savinov for feedback on the SPTM baseline. RS is supported by NSF grant IIS1763562, ONR grant N000141812861, AFRL CogDeCON, and Apple. Any opinions, findings and conclusions expressed in this material are those of the authors and do not necessarily reflect the views of NSF, AFRL, ONR, or Apple.
}
{\footnotesize

\bibliography{references}
\bibliographystyle{apalike}
}

\clearpage
\appendix

\section{Efficient Shortest Path Computation}
\label{appendix:graph-search}
   
\begin{wrapfigure}[17]{R}{0.55\textwidth}
\vspace{-2em}
\begin{minipage}[t]{0.55\textwidth}
  \begin{algorithm}[H]
  \caption{Inputs are the current state $s$, the goal state $g$, the replay buffer $\mathcal{B}$, and the value function $V$. Returns the length and first waypoint of the shortest path. \label{alg:shortest-path}}
  \begin{algorithmic}
   \Function{ShortestPath}{$s, s_g, \mathcal{B}, V$}
   \State // Matrices: $D_\pi, D_{\mathcal{B} \rightarrow \mathcal{B}}, D_{s \rightarrow s_g} \in \mathbb{R}^{|\mathcal{B}| \times |\mathcal{B}|}$
   \State // Vectors: $D_{s \rightarrow \mathcal{B}}, D_{\mathcal{B} \rightarrow g} \in \mathbb{R}^{|\mathcal{B}|}$
   \State $D_{\pi} \gets -V(\mathcal{B}, \mathcal{B})$ \Comment{cached}
   \State $D_{\mathcal{B} \rightarrow \mathcal{B}} \gets  \Call{FloydWarshall}{D_\pi}$ \Comment{cached}
   \State $D_{s \rightarrow \mathcal{B}} \gets -V(s, \mathcal{B})$
   \State $D_{\mathcal{B} \rightarrow g} \gets -V(\mathcal{B}, g)$
   \State $D_{s \rightarrow g} \gets D_{s \rightarrow \mathcal{B}} + D_{\mathcal{B} \rightarrow \mathcal{B}} + (D_{\mathcal{B} \rightarrow g})^T$
   \State $s_{w_1} \gets \displaystyle \arg\min_{u, v \in \mathcal{B}} D_{s \rightarrow g}$
   \State \Return $s_{w_1}$
   \EndFunction
    \end{algorithmic}
  \end{algorithm}
\end{minipage}
\end{wrapfigure}
Our policy solves a shortest path problem every time it recomputes a new waypoint. Na\"ively running Dijkstra's algorithm to compute a shortest path among the states in our active set $\mathcal{B}$ requires $O(|\mathcal{B}|^2)$ queries of our value function. While the search algorithm itself is fast, it is expensive to evaluate the value function on each pair of states at every time step.
In our implementation (Algorithm~\ref{alg:shortest-path}), we amortize this computation across many calls to the policy. We periodically periodically evaluate the value function on each pair of nodes in the replay buffer, and then used the Floyd Warshall algorithm to compute the shortest path between all pairs. This takes $O(|\mathcal{B}|^3)$ time, but only $O(|\mathcal{B}|^2)$ calls to the value function. Let $D \in \mathbb{R}^{|\mathcal{B}| \times |\mathcal{B}|}$ be the resulting matrix storing the shortest path distances between all pairs of states in the active set. Now, given a start state $s$ and goal state $g$, the shortest path distance is
\begin{equation*}
    d_{\text{sp}}(s, g) = \min \left( \min_{u, v \in \T} d(s, u) + D[u, v] + d(v, g), d(s, g)\right)
\end{equation*}
This computation requires $O(|\mathcal{B}|)$ calls to the value function, substantially better than the $O(|\mathcal{B}|^2)$ calls required with the na\"ive implementation.

\section{Environments}
\label{appendix:experimental-details}

We used two simple navigation environments, Point-U and Point-FourRooms, shown in Figure~\ref{fig:point-env}. In both environments, the observations are the location of the agent, $s = (x, y) \in \mathbb{R}^2$. The agent's actions $a = (dx, dy) \in [-1, 1]^2$ are added to the agents current position at every time step. We tuned the environments so that the goal-conditioned algorithm (which we will use as a baseline) would perform as well as possible. Observing that the agent would get stuck at corners, we modified the environment to automatically add Gaussian noise to the agents action. The resulting dynamics were
\begin{equation*}
    s_{t+1} = \texttt{proj}(s_t + a_t + \epsilon_t) \quad  \text{where} \quad \epsilon_t \sim \mathcal{N}(0, \sigma^2)
\end{equation*}
where \texttt{proj()} handles collisions with walls by projecting the state to the nearest free state. We used $\sigma^2 = 1.0$ for Point-U, and $\sigma^2=0.1$ for the (larger) Point-FourRooms environment.

\subsection{Visual Navigation}

We ran most experiments on SUNCG house \texttt{0bda523d58df2ce52d0a1d90ba21f95c}. We repeated all experiments on SUNCG house \texttt{0601a680273d980b791505cab993096a}, with nearly identical results. We manually choose houses using the following criteria (1) single story, (2) no humans, and (3) included multiple rooms to make planning challenging.
During training, we sampled ``nearby'' goal states (within 4 steps) for 80\% of episodes, and sampled goals uniformly at random for the remaining 20\% of episodes. We tuned these parameters to make goal-conditioned RL work as well as possible. We implemented goal-relabelling~\citep{kaelbling1993learning,andrychowicz2017hindsight}, choosing between the (1) originally sampled goal, the (2) current state, and (3) a future state in the same trajectory, each with probability 33\%.
The agent's actions space was to move North/South/East/West. Observations were panoramic images, created by concatenating the first-person views from each of the cardinal directions.
We used ensembles of 3 value functions, each with entirely independent weights.
For all neural networks conditioned on both the current observation and the goal observation, we concatenated the current observation and goal observation along their last channel. For RGB images, this resulted in an input with dimensions $H \times W \times 6$. For depth images, the concatenated input had dimension $H \times W \times 2$.

\section{Ablation Experiments}
\label{appendix:ablation}
\begin{figure}[H]
    \centering
    \begin{subfigure}[b]{0.45\textwidth}
    \includegraphics[width=\textwidth]{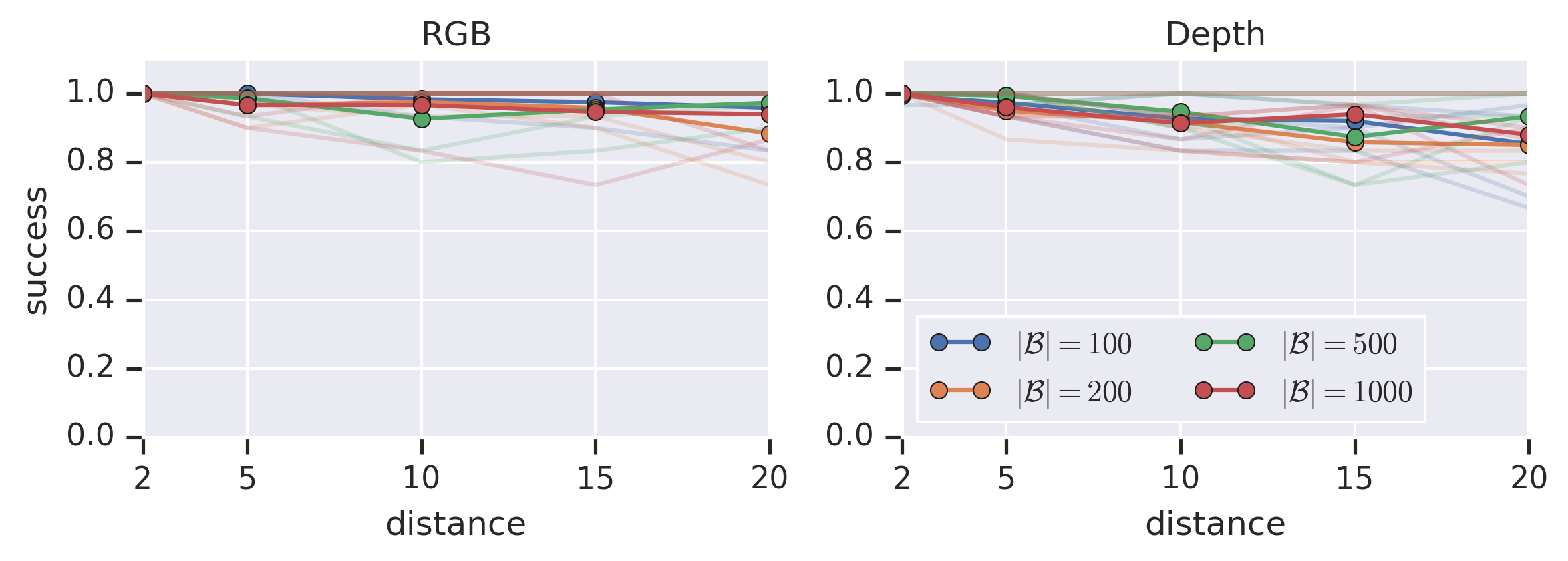}
    \caption{\textbf{Replay buffer size}~\label{fig:ablation-rb}}
    \end{subfigure}%
    ~
    \begin{subfigure}[b]{0.55\textwidth}
    \includegraphics[width=\textwidth]{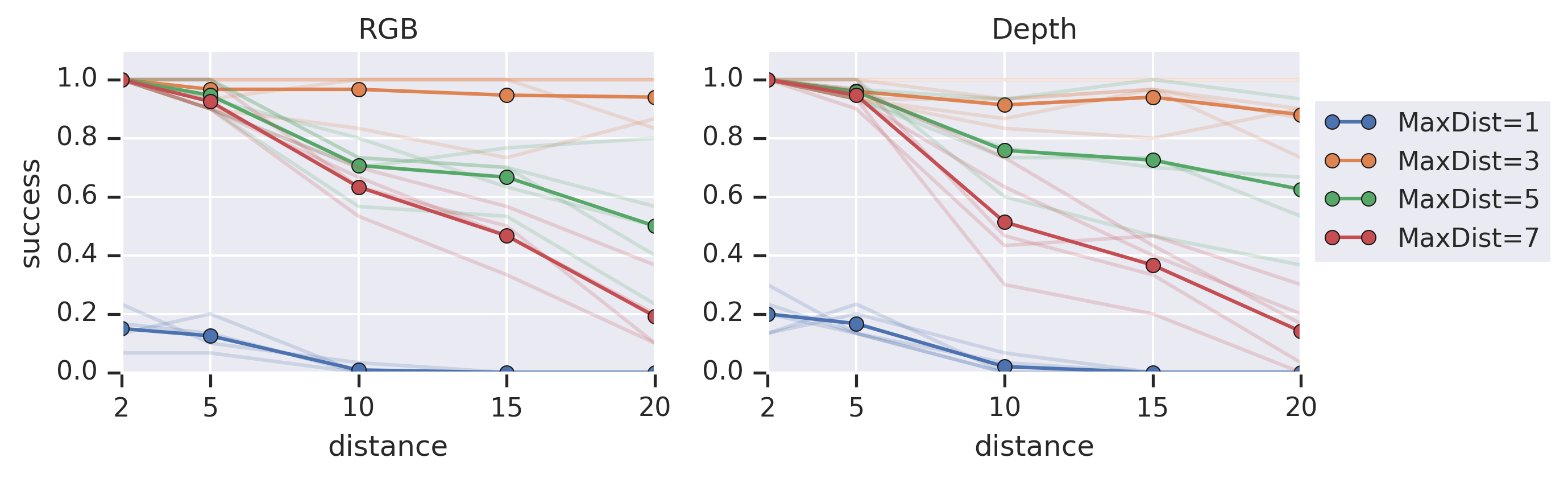}
    \caption{\textbf{Maximum edge length}~\label{fig:ablation-max-dist}}
    \end{subfigure}
    \caption{\textbf{Sensitivity to Hyperparameters}: \figleft\; While we used a buffer of 1000 observations for most of our experiments, decreasing the buffer size has little effect on the method's success rate. \figright\; When constructing our graph, we ignore edges that are longer than some distance, \textsc{MaxDist}. We find that this hyperparameter is important to the success of our method.
    \label{fig:ablation-1}}
\end{figure}
Because SoRB plans over a fixed replay buffer, one potential concern is that performance might degrade if the replay buffer is too small. To test this concern, we ran an experiment varying the size of the replay buffer. 
As shown in Figure~\ref{fig:ablation-rb}, decreasing the replay buffer by a factor of 10x led to no discernible drop on performance.
While we do expect performance to drop if we further decrease the size of the replay buffer, the requirement of storing 100 states (even high-resolution images) seems relatively minor. In a second ablation experiment, we varied the \textsc{MaxDist} hyperparameter that governs when we stop adding new edges to the graph. As shown in Figure~\ref{fig:ablation-max-dist}, SoRB is sensitive to this hyperparameter, with values too large and too smaller leading to worse performance. When the \textsc{MaxDist} parameter is too small, graph search fails to find a path to the goal state. As we increase \textsc{MaxDist}, we increase the probability of underestimating the distance between pairs of states. We expect that improvements in uncertainty quantification in RL will improve the stability of our method w.r.t. this hyperparameter.

\section{Tricks for Learning Distances with RL}
\begin{enumerate}
    \item \emph{Small learning rates}: Especially for the image-based tasks, we found that RL completely failed with using a critic learning rate larger than 1e-4. Smaller learning rates work too, but take longer to converge.
    \item \emph{Distributional RL}: The value function update for distributional RL has a particularly nice form when values correspond to distances. Additionally, distributional RL implicitly clips the values, preventing the critic to predict that unreachable states are infinitely far away.
    \item \emph{Termination Condition}: Carefully consider whether to set \texttt{done = True} at the end of each episode. In our setting the agent received a reward of -1 at each time step, so the value of each state was negative. An optimal agent therefore attempts to terminate the episode as quickly as possible. We only set \texttt{done = True} when the agent reached the goal state, not when the maximum number of time steps was reached or when it reached some other absorbing state.
    \item \emph{Ensembles of Value Functions}: Predicted distances from a single value function can be inaccurate for unseen (state, goal) pairs. When performing search using these predicted distances, these inaccurately-short predictions result in ``wormholes'' through the environment, where the agent mistakenly believes that two distant states are actually nearby. To mitigate this, we trained multiple, independent critics in parallel on the same data, and then aggregated predictions from each before doing search. Surprisingly, we found that taking the average predicted distance over the ensemble worked as well as taking the maximum predicted distance. We tried accelerating training by using shared convolutional layers for all critics in the ensemble, but found that this resulted in highly-correlated distant predictions that exhibited the ``wormhole'' problem.
    \item \emph{Normalizing Observations}: For the visual navigation experiments, we normalized the observations to be in the interval $[0, 1]$ by dividing by the maximum pixel intensity (32 for depth, 255 for RGB). Normalization was most important for the generalization experiment with RGB observations.
\end{enumerate}

\section{Failed Experiments}
\begin{enumerate}
    \item \emph{Goal Relabelling}: As mentioned above, we tried to combine our method with off-policy goal relabelling~\citep{andrychowicz2017hindsight,pong2018temporal}. Surprisingly, we found that this hurt performance of the non-search policy, and had no effect on the search policy.
    \item \emph{Lower-bounds on Q-values}: We attempted to use the search path to obtain a lower bound on the target Q-values during training. In the Bellman update, we replaced the distance predicted by the target Q-values with the minimum of (1) the distance predicted by the target Q-network and (2) the distance of the shortest path found by search. This can be interpreted as a generalization of the single-step lower bound from~\citet{kaelbling1993learning}. Initial experiments showed this approach slowed down learning, and in some cases prevented the algorithm from converging. We hypothesize that Q-learning is must more sensitive to error in the \emph{relative values} of two actions, rather than the \emph{absolute value} of any particular action. While our lower-bound method likely decreased the absolute error, it did not decrease the relative error (and may have even increased it).
    \item \emph{TD3-style Ensemble Aggregation}: In our main experiments, we aggregated distance predictions from the ensemble of distributional critics by first computing the expected distance of each critic, and then taking the maximum predicted distance. This approach ignores the fact that our critics are distributional. Inspired by the stability of TD3, we attempted to apply a similar approach to aggregating predictions from the ensemble of distributional critics. The na\"ive approach of taking the minimum for each atom does not work because the resulting distribution will not sum to one. Instead, we first compute the cumulative density function (CDF) of each critic and then take the pointwise maximum over the CDFs. Note that critics correspond to negative distance, so the maximum corresponds to being pessimistic. Finally, we convert the resulting CDF back into a PDF and return the corresponding expected distance. While this method has neat connections to second-order stochastic dominance and risk-averse expected utility maximizers~\citep{hadarUncertain}, we found that it worked poorly in practice.
\end{enumerate}

\section{Hyperparameters}
\label{appendix:hparams}
Unless otherwise noted, all baselines use the same hyperparameters as our method. Unless otherwise noted, parameters were not tuned.

\subsection{Search on the Replay Buffer}

\begin{table}[H]
    \centering
    \tiny
    \begin{tabular}{p{3cm}|p{4cm}|p{4cm}}
        \textbf{Parameter} & \textbf{Value}  & \textbf{Comments} \\
        \hline
        learning rate &  1e-4 & Lower values also work, but training takes longer. Same for actor and critic. \\
        \hline
        training iterations & 1e6 environment steps & Performance changed little after 200k steps. \\
        \hline
        batch size & 64\\
        \hline
        train steps per environment step & 1:1\\
        \hline
        random steps at start of training & 1000\\
        \hline
        NN architecture (images) & Conv(16, 8, 4) + Conv(32, 4, 4) + FC(256) & Same for depth and RGB images. \\
        \hline
        optimizer & Adam & We used the default Tensorflow settings for $\beta_1, \beta_2, \epsilon$. Same for actor and critic.\\
        \hline
        MaxDist & 3 & See Figure~\ref{fig:ablation-1}\\
        \hline
        replay buffer size (training) & 100k \\
        \hline
        replay buffer size (search) & 1k & See Figure~\ref{fig:ablation-1}\\
        \hline
        gamma / discount & 1 \\
        \hline
        $\epsilon$  & 0.1 & Exploration parameter for discrete actions, used for visual navigation. \\
        \hline
        OU-stddev, OU-damping & 1.0, 2.0 & Exploration parameters for continuous actions, used for didactic 2D navigation \\
        \hline
        reward scale factor & 0.1 & Tuned for the DDPG baseline on the 2D navigation task. \\
        \hline
        target network update frequency & every 5 steps \\
        \hline
        target network update rate ($\tau$) & 0.05 \\
        \hline
    \end{tabular}
    \caption{Hyperparameters for SoRB}
\end{table}

\subsection{Value Iteration Networks}

\begin{table}[H]
    \tiny
    \centering
    \begin{tabular}{p{3cm}|p{4cm}|p{4cm}}
        \textbf{Parameter} & \textbf{Value}  & \textbf{Comments} \\
        \hline
        number of iterations & 50 & Tuned over [1, 2, 5, 10, 20, 50]. Little effect. \\
        \hline
        hidden units in VI block & 100 & Tuned over [10, 30, 100, 300]. Little effect\\
        \hline
    \end{tabular}
    \caption{Hyperparameters for VIN~\citep{tamar2016value}}
    \label{fig:hparams}
\end{table}

\subsection{Semi-Parametric Topological Memory}
We first tuned the $l$ parameter on goal-reaching without search. Setting $l$ to the best found value, we performed a massive (over 1000 experiments) grid search over $M$, $s_{\text{reach}}$, and the threshold for adding edges.
\begin{table}[H]
    \tiny
    \centering
    \begin{tabular}{p{3cm}|p{4cm}|p{4cm}}
        \textbf{Parameter} & \textbf{Value}  & \textbf{Comments} \\
        \hline
        threshold for adding edges & 0.9 & Tuned over [0.1, 0.2, 0.5, 0.7, 0.9] \\
        \hline
        $s_{\text{reach}}$, threshold for choosing the next waypoint along the shortest path & 0.5 & Tuned over [0.0, 0.1, 0.2, 0.3, 0.4, 0.5, 0.6, 0.7, 0.8, 0.9, 0.95, 1.0]\\
        \hline
        NN architecture & Conv(16, 8, 4) + Conv(32, 4, 4) + FC(256) & Same architecture (but different weights) for the retrival and locomotor networks. \\
        \hline
        $l$, threshold for sampling nearby states in trajectory & 8 & Tuned over [1, 2, 4, 8] \\
        \hline
        $M$, margin between ``close'' and ``far'' states & 1 & Tuned over [1, 2, 4] \\
        \hline
    \end{tabular}
    \caption{Hyperparameters for SPTM~\citep{savinov2018semi}}
    \label{fig:hparams}
\end{table}

\end{document}